\documentclass[conference]{IEEEtran}
\usepackage[usenames,dvipsnames]{color}
\usepackage[latin1]{inputenc}
\usepackage[labelfont=bf]{caption}
\usepackage{enumerate}
\usepackage[colorlinks=true,urlcolor=blue,linkcolor=black,citecolor=blue,bookmarks=true]{hyperref}
\usepackage{graphicx}
\usepackage{amsmath}
\usepackage{amssymb}
\usepackage{balance}
\usepackage{xspace}
\usepackage{multirow}
\usepackage{multicol}

\usepackage{microtype}
\DisableLigatures[f]{encoding = *, family = * }

\graphicspath{{./}{./figures/}}

\def\Vec#1{{\boldsymbol{#1}}}
\def\Mat#1{{\boldsymbol{#1}}}

\def\ie{ie.\xspace}

\begin{document}

\title{\huge Bags of Affine Subspaces for Robust Object Tracking}

\author
  {
  Sareh Shirazi$^{{\tiny ~}\dagger\ddagger}$, Conrad Sanderson$^{{\tiny ~}\circ\ast}$, Chris McCool$^{{\tiny ~}\ddagger}$, Mehrtash T. Harandi$^{{\tiny ~}\circ\triangledown}$\\
  ~\\
  $^{\dagger{\tiny ~}}$Australian Centre for Robotic Vision (ACRV)\\
  $^{\ddagger{\tiny ~}}$Queensland University of Technology, Australia\\
  $^{\triangledown{\tiny ~}}$Australian National University, Australia\\
  $^{\ast{\tiny ~}}$University of Queensland, Australia\\
  $^{\circ{\tiny ~}}$NICTA, Australia\\
  ~
  }
  
\maketitle

\begin{abstract}

We propose an adaptive tracking algorithm where the object is modelled as a continuously updated bag of affine subspaces,
with each subspace constructed from the object's appearance over several consecutive frames.
In contrast to linear subspaces, affine subspaces explicitly model the origin of subspaces.
Furthermore, instead of using a brittle point-to-subspace distance during the search for the object in a new frame,
we propose to use a subspace-to-subspace distance by representing candidate image areas also as affine subspaces.
Distances between subspaces are then obtained by exploiting the non-Euclidean geometry of Grassmann manifolds.
Experiments on challenging videos (containing object occlusions, deformations, as well as variations in pose and illumination)
indicate that the proposed method
achieves higher tracking accuracy than several recent discriminative trackers.
\end{abstract}

\section{Introduction}
\label{sec:introduction}

Object tracking is a core task in applications such as automated surveillance, traffic monitoring and human behaviour analysis~\cite{liu2013intelligent,yilmaz2006object}.
Tracking algorithms need to be robust to intrinsic object variations (eg., shape deformation and pose changes)
and extrinsic variations (eg., camera motion, occlusion and illumination changes)~\cite{yilmaz2006object}.

In general, tracking algorithms can be categorised into two main categories:
{\bf (i)}~generative tracking~\cite{adam2006robust,ross2008incremental,Wang2013},
and
{\bf (ii)}~discriminative tracking~\cite{babenko2011,kalal2011tracking,lu2012pixel}.
Generative methods represent the object as a particular appearance model
and then focus on searching for the location that has the most similar appearance to the object model.
Discriminative approaches treat tracking as a binary classification task,
where a discriminative classifier is trained to explicitly separate the object from non-object areas such as the background.
To achieve good performance, discriminative methods in general require a larger training dataset than generative methods.

A promising approach for generative tracking is to model object appearance via subspaces
\cite{ho2004visual,li2004incremental,ross2008incremental,Yang_TIP_2009}.
A~common approach in such trackers is to apply eigen-decomposition on a set of object images,
with the resulting eigenvectors defining a linear subspace.
These linear subspaces are able to capture perturbations of object appearance due to variations in
viewpoint,
illumination,
spatial transformation,
and articulation.
However, there are two major shortcomings.
First, a linear subspace does not model the mean of the image set (ie., origin of the subspace) which can potentially hold useful discriminatory information;
all linear subspaces have a common origin.
Second, subspace based trackers typically search for the object location by comparing candidate image areas to the object model (linear subspace) using a brittle point-to-subspace distance~\cite{li2007robust,skocaj2003}
(also known as distance-from-feature-space~\cite{wang2008manifold}),
which can be easily affected by drastic appearance changes such as partial occlusions.
For face recognition and clustering it has been shown that 
improved performance can be achieved when subspace-to-subspace distances are used instead~\cite{basri2011approximate,fitzgibbon2003joint,Sanderson_AVSS_2012}.

To address the shortcomings of traditional subspace based trackers,
in this work%
\footnote{This paper is a thoroughly revised and extended version of our earlier preliminary work~\cite{Shirazi_WACV_2014}.}
we propose a tracker with the following four characteristics:

\begin{enumerate}[{\bf (1)~}]
\renewcommand{\itemsep}{1ex}

\item
Instead of linear subspaces, we propose to model object appearance using affine subspaces, thereby taking into account the origin of each subspace.

\item
Instead of using point-to-subspace distance,
we propose to represent the candidate areas as affine subspaces
and use a subspace-to-subspace distance;
this allows for more robust modelling of the candidate areas and in effect increases the memory of the tracker.

\item
To accurately measure distances between subspaces,
we exploit the non-Euclidean geometry of Grassmann manifolds~\cite{Harandi_2013_PRL,Lui_IVC_2012,Sanderson_AVSS_2012}. 

\item
To take into account drastic appearance changes that are not well modelled by individual subspaces (such as occlusions)~\cite{yang2011},
the tracked object is represented by a continuously updated bag of affine subspaces;
this is partly inspired by~\cite{babenko2011}, where bags of object images are used.

\end{enumerate}

~

To the best of our knowledge,
this is the first time that appearance is modelled by affine subspaces for object tracking.
The proposed approach is somewhat related to adaptive subspace tracking~\cite{ho2004visual,ross2008incremental,wang2008online}.
In~\cite{ho2004visual,ross2008incremental} an object is represented as a single low-dimensional linear subspace,
which is constantly updated using recent tracking results.
In~\cite{wang2008online}, an online subspace learning scheme employing Grassmann manifolds is used to update the object model.
In the above methods, only linear subspaces and point-to-subspace distances are considered.
In contrast, the proposed method uses affine subspaces and a more robust subspace-to-subspace distance.
Furthermore, instead of updating a single subspace, the proposed method keeps a bag of recent affine subspaces,
where old subspaces are replaced with new ones.

We continue the paper as follows.
An overview of related work is given in Section~\ref{sec:prior_work}.
Section~\ref{sec:model} presents the proposed tracking approach in detail.
Comparative evaluations against several recent tracking methods are reported in Section~\ref{sec:experiments}.
The main findings and possible future directions are given in~Section~\ref{sec:conclusion}.

\section{Related Work}
\label{sec:prior_work}

In this section, we first overview the evolution of subspace-based trackers.
We then briefly describe two popular generative trackers:
the mean shift tracker~\cite{comaniciu2003} and the fragments-based tracker~\cite{adam2006robust}.
Finally, we briefly cover two recent discriminative tracking methods:
Multiple Instance Learning (MIL) tracker~\cite{babenko2011} and Tracking-Learning-Detection (TLD)~\cite{kalal2011tracking}.

\subsection{Subspace Based Trackers}

As the main challenge in visual tracking is the difficulty in handling the appearance variability of a target object,
it is imperative for a robust tracking algorithm to model such appearance variations.
This can be difficult to accomplish when the object model is based on only a single image.
Subspaces allow us to group images together and provide a single representation as a compact appearance model~\cite{ross2008incremental}. 
Subspace-based tracking originated with the work of Black and Jepson~\cite{black1998},
where a subspace learning-based approach is proposed for tracking rigid and articulated objects.
This approach uses a view-based eigenbasis representation with parameterised optical flow estimation.
As the algorithm is based on iterative parameterised matching between the eigenspace and candidate image regions,
it might have a relatively high computational load~\cite{la2000fast}.
It also uses a single pre-trained subspace to provide the object appearance model across the entire video.
As such, to achieve robust visual tracking with this method,
it is necessary to first collect a large set of training images covering the range of possible appearance variations,
which can be difficult to accomplish in practice.

Addressing the limitations of having a single representation for object appearance which is always learned off-line before tracking begins,
Skocaj and Leonardis~\cite{skocaj2003} developed a weighted incremental Principal Component Analysis (PCA) approach for sequentially updating the subspace.
Although the method improves tracking accuracy,
it has the limitation of being computationally intensive due to an optimisation problem that has to be computed iteratively.
To address this issue, Li et al.~\cite{li2004incremental} proposed an alternative incremental PCA-based algorithm for subspace learning.
In this approach, the PCA model updating is performed directly using the previous eigenvectors and a new observation vector,
thereby significantly decreasing the computational load of the update process.

Ho et al.~\cite{ho2004visual} proposed an adaptive tracker using a uniform $L_2$-reconstruction error norm for subspace estimation,
allowing explicit control on the approximation quality of the subspace.
Empirical results show increases in tracking robustness and more swift reactions to environmental changes.
However, as the method represents objects as a point in a linear subspace computed using only recent tracking results,
the tracker may drift if large appearance changes occur~\cite{hu2012single}.

Lim et al.~\cite{lim2004incremental} proposed a generalised tracking framework which constantly learns and updates a low dimensional subspace representation of the object.
The updates are done using several observations at a time instead of a single observation.
To estimate the object locations in consecutive frames, a sampling algorithm is used with robust likelihood estimates.
The likelihood for each observed image being generated from a subspace is inversely proportional to the distance of that observation from the subspace.
Ross et al.~\cite{ross2008incremental} improved the tracking framework in~\cite{lim2004incremental}
by adding a forgetting factor to focus more on recently acquired images and less on earlier observations during the learning and update stages. 

Hu et al.~\cite{hu2012single} presented an incremental log-Euclidean Riemannian subspace learning algorithm
in which covariance matrices of image features are mapped from a Riemannian manifold into a vector space,
followed by linear subspace analysis.
A block based appearance model is used to capture both global and local spatial layout information.
Similar to traditional subspace based trackers, this method also uses a point-to-subspace distance.

\subsection{Other Generative Trackers}

Among algorithms that do not use subspaces,
two popular generative trackers are the mean shift tracker~\cite{comaniciu2003} and the fragments-based tracker~\cite{adam2006robust}.
The mean shift tracker models object appearance with colour histograms which can be applied to track non-rigid objects. 
Both the object model and candidate image areas are represented by colour pdfs,
with the Bhattacharyya coefficient used as the similarity measure~\cite{kailath1967}.
Tracking is accomplished by finding the local maxima of the similarity function using gradient information provided by the mean shift vector which always points toward the direction of maximum.  
While effective, the mean shift tracker is subject to several issues.
First, the spatial information is lost, which precludes the application of more general motion models~\cite{adam2006robust,yang2005efficient}. 
Second, the Bhattacharyya coefficient may not be discriminative enough for tracking purposes~\cite{yang2005efficient}.
Third, the method only maintains a single template to represent the object,
leading to accuracy degradation if an object moves rapidly or if a significant occlusion occurs.

The fragments-based tracker~\cite{adam2006robust} aims to handle partial occlusions via a parts-based model. 
The object is represented by multiple image fragments or patches. 
Spatial information is retained due to the use of spatial relationships between patches.
Each patch votes on the possible positions and scales of the object in the current frame,
by comparing its histogram with histograms of image patches in the frame.
The tracking task is carried out by combining the vote maps of multiple patches by minimising a robust statistic.
However, the object model is not updated and thereby it is not expected to handle tracking objects that exhibit significant appearance changes~\cite{wang2011superpixel,babenko2011}.

\subsection{Discriminative Trackers}

Two recent discriminative methods are 
the Multiple Instance Learning tracker (MILTrack)~\cite{babenko2011}
and the Tracking-Learning-Detection~(TLD) approach~\cite{kalal2011tracking}.
In the MILTrack approach, 
instead of using a single positive image patch to update the classifier,
a set of positive image patches is maintained and used to update a multiple instance learning classifier~\cite{dietterich1997}.
In multiple instance learning, training examples are presented in sets with class labels provided for entire sets rather than individual samples.
The use of sets of images allows the MILTrack approach to achieve robustness to occlusions and other appearance changes.
However, if the object location detected by the current classifier is imprecise,
it may lead to a noisy positive sample and consequently a suboptimal classifier update. 
These noisy samples can accumulate and cause tracking drift or failure~\cite{zhang2013real}.

The TLD approach decomposes the tracking task into three separate tasks: tracking, learning and detection. 
It regards tracking results as unlabelled and exploits their underlying structure
using positive and negative experts to select positive and negative samples for update.
This method makes a common assumption in tracking that the training samples follow the same distribution as the candidate samples.
Such an assumption is problematic if the object's appearance or background changes drastically or continuously,
which causes the underlying data distribution to keep changing~\cite{li2013ssocbt}.

\section{Proposed Tracking Approach}
\label{sec:model}

The proposed tracking approach is comprised of four intertwined components, listed below.
To ease understanding of the overall system, we first overview the components below,
and then provide the details for each component in the following subsections.

\begin{enumerate}[{\it(A)~}]
\renewcommand{\itemsep}{1ex}

\item
{\it Particle Filtering Framework}.
An object's location in consecutive frames is parameterised as a distribution in a particle filter framework~\cite{maskell2001tutorial},
where a set of particles represents the distribution and each particle represents a location.
The location history of the tracked object in previous frames is taken into account to create a set of candidate object locations in a new frame.

\item
{\it Particle Representation}.
We represent the $i$-th particle at time~$t$ using an affine subspace {$\mathcal{A}_{i}^{(t)}$},
which is constructed by taking into account the appearance of the $i$-th candidate location at time~$t$ as well as 
the appearance of the tracked object in several immediately preceding frames.
Each affine subspace {$\mathcal{A}_{i}^{(t)}$}
is comprised of mean {$\Vec{\mu}_{i}^{(t)}$} and basis~{$\Mat{U}_{i}^{(t)}$}.

\item
{\it Bag of Affine Subspaces}.
To take into account drastic appearance changes, the tracked object is modelled by a set of affine subspaces,
which we refer to as bag {$\mathcal{B}$}.
During tracking the bag first grows to a pre-defined size, 
and then its size is kept fixed by replacing the oldest affine subspace with the latest affine subspace.

\item
{\it Comparing Affine Subspaces}.
Each candidate subspace {$\mathcal{A}_{i}^{(t)}$} from the pool of candidates is compared to the affine spaces in bag {$\mathcal{B}$}.
The most likely candidate subspace is deemed to represent the best particle,
which in turn indicates the new location of the tracked object.
The distance between affine subspaces is comprised of the distance between their means
and the Grassmann geodesic distance between their bases.

\end{enumerate}

\subsection{Particle Filtering Framework}
\label{sec:particle_filtering_framework}

We aim to obtain the location {$x \in \mathcal{X}$, $y \in \mathcal{Y}$}
and the scale {$s \in \mathcal{S}$} of an object in frame~$t$ based on information obtained from previous frames.
A~blind search in the space of location and scale is inefficient,
since not all possible combinations of $x$, $y$ and~$s$ are plausible.
To efficiently search the location and scale space,
we adapt a particle filtering framework~\cite{maskell2001tutorial,yilmaz2006object},
where the object's location in consecutive frames is parameterised as a distribution.
The distribution is represented using a set of particles, with each particle representing a location and scale.

Let {$\Vec{z}_i^{(t)} = [ x_i^{(t)}, y_i^{(t)}, s_i^{(t)} ]^T$} denote the state of the $i$-th particle comprised of the location and scale at time~$t$.
Using importance sampling~\cite{maskell2001tutorial},
the density of the location and scale space (or most probable candidates) at time~$t$ is estimated as a set of $N$ particles
{$\{\Vec{z}_{i}^{(t)}\}_{i=1}^{N}$} using particles from the previous frame {$\{\Vec{z}^{(t-1)}_{i}\}_{i=1}^{N}$} and
their associated weights \mbox{\small{$\{w^{(t-1)}_{i}\}_{i=1}^{N}$}}
(with constraints {$\sum\nolimits_{i=1}^N w^{(t-1)}_{i} = 1$} and each $w_i\geq 0$).
For now we assume the associated weights of particles are known and later discuss how they can be determined.

To generate {$\{\Vec{z}_{i}^{(t)}\}_{i=1}^{N}$},
{$\{\Vec{z}^{(t-1)}_{i}\}_{i=1}^{N}$} is first sampled (with replacement) $N$ times.
The probability of choosing {$\Vec{z}^{(t-1)}_{i}$}, the $i$-th particle at time $t-1$, is equal to the associated weight {$w^{(t-1)}_{i}$}.
Each chosen particle then undergoes an independent Brownian motion, which is modelled by a Gaussian distribution.
As a result, for a chosen particle {$\Vec{z}^{(t-1)}_{i}$}, a new particle {$\Vec{z}^{(t)}_{i}$}
is obtained as a random sample from {$\mathcal{N}( \Vec{z}^{(t-1)}_i, \Mat{\Sigma} )$},
where {$\mathcal{N}(\Vec{\mu}, \Mat{\Sigma})$}
denotes a Gaussian distribution with mean~{$\Vec{\mu}$} and diagonal covariance matrix~{$\Mat{\Sigma}$}.
The latter governs the speed of motion by controlling the location and scale variances.

\subsection{Particle Representation via Affine Subspaces}
\label{sec:particle_representation}

To accommodate a degree of variations in object appearance,
particle {$\Vec{z}_i^{(t)}$} is represented by an affine subspace~{$\mathcal{A}_{i}^{(t)}$},
constructed from the appearance of the $i$-th candidate location at time~$t$ as well as 
the appearance of the tracked object in several immediately preceding frames.
Each affine subspace {$\mathcal{A}_{i}^{(t)}$} can be described by a 2-tuple:
\begin{equation}
    \mathcal{A}_{i}^{(t)} = \left \{ \Vec{\mu}_{i}^{(t)} , \Mat{U}_{i}^{(t)} \right \}
    \label{eqn:raff}
\end{equation}
\noindent
where
{$\Vec{\mu}_{i}^{(t)} \in \mathbb{R}^{D}$}
is the origin (mean) of the subspace
and
{$\Mat{U}_{i}^{(t)} \in \mathbb{R}^{D \times n}$}
is the basis of the subspace.
The parameter $n$ is the number of basis vectors.

The subspace is obtained as follows.
Let {$\Vec{v}(\Vec{z}_{i}^{(t)})$} represent the vectorised form of the $i$-th candidate image patch at time~$t$.
The top-left corner of the patch is indicated by ({$x_{i}^{(t)},y_{i}^{(t)}$}) and its size by {$s_{i}^{(t)}$}.
The patch is resized to a fixed size of {$H_{1} \times H_{2}$} pixels
and represented as a column vector of size {$D = {H_{1} \times H_{2}}$}.
In the same manner, 
let {$\Vec{v}(\Vec{z}_{\ast}^{(t-1)})$} denote the vectorised form of 
the appearance of the tracked object at time {$(t-1)$},
with {$\Vec{z}_{\ast}^{(t-1)}$} denoting the particle that was deemed at time {$(t-1)$} to represent the tracked object.
The vectorised forms of the candidate image patch as well as the patches containing the tracked object in the previous~{$P$} frames
are used to construct the following {$D \times {(P+1)}$} sized matrix:
\begin{equation}
  \Mat{V}_{i}^{(t)} = \left[ \Vec{v}(\Vec{z}_{i}^{(t)}), \Vec{v}(\Vec{z}_{\ast}^{(t-1)}), \cdots, \Vec{v}(\Vec{z}_{\ast}^{(t-P)})   \right]
  \label{eqn:vectorised_patch_collection}
\end{equation}
\noindent
The subspace origin $\Vec{\mu}_{i}^{(t)}$ is the mean of {$\Mat{V}_{i}^{(t)}$}.
The subspace basis {$\Mat{U}_{i}^{(t)}$} is obtained by performing singular value decomposition (SVD) of
{$\Mat{V}_{i}^{(t)}$} and choosing the $n$ dominant left eigenvectors corresponding to the $n$ largest eigenvalues.

\subsection{Bag of Affine Subspaces}
\label{Bag_Affine_Subspaces}

To take into account drastic appearance changes that might not be well modelled by subspaces,
we propose to adapt the approach of keeping a history of object appearance variations~\cite{babenko2011}, 
by modelling the tracked object via a set of affine subspaces obtained during the tracking process.
We refer to such a set as a {\it bag} of affine subspaces, defined as:
\begin{equation}
  \mathcal{B} = \{\mathcal{A}_{1}, \cdots, \mathcal{A}_{K}\}
\end{equation}
\noindent
where {$K$} is the number of subspaces to keep.
The bag is updated every $W$ frames by replacing the oldest affine subspace with the latest.
The size of bag determines the memory of the tracking system.

To demonstrate the benefit of the bag approach, consider the following scenario.
A~person is being tracked, with the appearance of their whole body modelled as a single subspace.
At some point a partial occlusion occurs, and only the upper body is visible for several frames.
The tracker then learns the new occluded appearance of the person.
If the tracker is only aware of the very last seen appearance (ie.,~the upper body),
the tracker is likely to lose the object upon termination of the occlusion.
Keeping a set of affine subspaces (ie.,~both upper body and whole body)
increases memory of the tracked object
and hence can help to overcome the confounding effect of drastic appearance changes.

\subsection{Comparing Affine Subspaces}
\label{Comparing_Affine_Subspaces}

Each candidate subspace {$\mathcal{A}_{i}^{(t)}$} from the pool of candidates is compared to the affine spaces in bag {$\mathcal{B}$}.
The most likely candidate subspace is deemed to represent the best particle,
which in turn indicates the new location and scale of the tracked object.

The simplest distance measure between two affine subspaces is the minimal Euclidean distance, 
ie.,~the minimum distance of any pair of points of the two subspaces.
However, such a measure does not form a metric~\cite{basri2011approximate}
and it does not consider the angular distance between affine subspaces, which can be a useful discriminator~\cite{kim2007discriminative}.
On the other hand, using only the angular distance 
ignores the origin of affine subspaces and reduces the problem to a linear subspace case,
which we wish to avoid.

To address the above limitations, we propose a distance measure with the following form:
\begin{equation}
    \operatorname{dist}(\mathcal{A}_{i}, \mathcal{A}_{j})
    =
    \alpha~\widehat{\operatorname{d}}_{o} \left(\Vec{\mu}_{i},\Vec{\mu}_{j}\right)
    +
    (1-\alpha)~\widehat{\operatorname{d}}_g \left(\Mat{U}_{i},\Mat{U}_{j}\right)
    \label{eqn:r2}
\end{equation}%
\noindent
where $\alpha \in [0,1]$ is a mixing weight,
while
{$\widehat{\operatorname{d}}_o(\cdot,\cdot) \in [0,1]$} 
is a normalised distance between the origins of the subspaces
and {$\widehat{\operatorname{d}}_g(\cdot,\cdot) \in [0,1]$} 
is a normalised Grassmann geodesic distance between bases of the subspaces.

We define the distance between the origins of $\mathcal{A}_{i}$ and $\mathcal{A}_{j}$~as:
\begin{equation}
  \widehat{\operatorname{d}}_{o} \left(\Vec{\mu}_{i},\Vec{\mu}_{j}\right) = \gamma \| \Vec{\mu}_{i} - \Vec{\mu}_{j} \|^{2}
  \label{eqn:normalised_origin_dist}
\end{equation}
\noindent
where $\gamma$ is a scaling parameter.
Under the assumption that normalised images are used so that each pixel value is in the $[0,1]$ interval,
the elements of $\Vec{\mu} \in \mathbb{R}^D$ are also in the $[0,1]$ interval.
As such, the maximum value of the $\| \Vec{\mu}_{i} - \Vec{\mu}_{j} \|^{2}$ component in Eqn.~(\ref{eqn:normalised_origin_dist})
is equal to~$D$, and hence $\gamma = 1/D$.

A Grassmann manifold (a special type of Riemannian manifold)
is defined as the space of all $n$-dimensional linear subspaces of {$\mathbb{R}^D$} for \mbox{$0<n<D$}~\cite{Absil_2008,edelman1998geometry,Harandi_IJCV_inpress,Harandi_2013_PRL,Lui_IVC_2012}.
A point on Grassmann manifold {$\mathcal{G}_{D,n}$}
is represented by an orthonormal basis through a \mbox{$D \times n$} matrix.
The length of the shortest smooth curve connecting two points on a manifold is known as the geodesic distance.
For Grassmann manifolds, the squared geodesic distance between subspaces
{$\Mat{E}$} and {$\Mat{F}$} is given by:
\begin{equation}
    \operatorname{d}_g\left(\Mat{E},\Mat{F}\right) = \| \Theta \|^{2}
    \label{eqn:geodesic_Grass}
\end{equation}%
\noindent
where {$\Theta=[\theta_1,\theta_2,\cdots,\theta_n]$} is the principal angle vector, \ie
\begin{equation}
  \cos(\theta_k)
  =
  \max_{\Vec{e}_k^{~} \in \Mat{E}, ~ \Vec{f}_k^{~} \in \Mat{F}}
  \Vec{e}_k^T\Vec{f}_k^{~}
  \label{eqn:Principal_Angle}
\end{equation}%
\noindent
subject to
\mbox{$\left \| \Vec{e}_k^{~} \right \| \mbox{~=~} \left \| \Vec{f}_k^{~} \right \| \mbox{~=~} 1$},
\mbox{$\Vec{e}_k^T \Vec{e}_l^{~} \mbox{~=~} \Vec{f}_k^T \Vec{f}_l^{~} \mbox{~=~} 0$},
\mbox{$l \mbox{~=~} 1, \ldots, k\mbox{--}1$}.
In other words, the first principal angle $\theta_1$ is the smallest angle between all pairs of unit vectors in the two subspaces,
with the remaining principal angles defined similarly.
The principal angles can be computed through the SVD of \mbox{$\Mat{E}^T \Mat{F}$},
with the $k$-th singular value corresponding to $\cos(\theta_k)$~\cite{edelman1998geometry,Absil_2008}.
The principal angles have the property of \mbox{$\theta_i \in [0, \pi/2]$}.
As such, the maximum value of {$\operatorname{d}_g\left(\Mat{E},\Mat{F}\right)$ is $n \pi^2/4$}.
Therefore, we define the normalised squared Grassmann geodesic distance as
\begin{equation}
  \widehat{\operatorname{d}}_g\left(\Mat{E},\Mat{F}\right) = \beta \operatorname{d}_g\left(\Mat{E},\Mat{F}\right)
  \label{eqn:geodesic2}
\end{equation}%
where {$\beta = 4/(n\pi^2)$}.

To measure the overall likelihood of a candidate affine subspace {$\mathcal{A}_{i}^{(t)}$} according to bag {$\mathcal{B}$},
the individual likelihoods of {$\mathcal{A}_{i}^{(t)}$} according to each affine subspace in {$\mathcal{B}$}
are integrated using a straightforward sum rule~\cite{kittler1998combining,Sanderson_DSP_2004}:
\begin{equation}
    p \left( \mathcal{A}_{i}^{(t)}|\mathcal{B} \right)
    =
    \sum\nolimits_{k=1}^{K} {\widehat{p} \left( \mathcal{A}_{i}^{(t)}|\mathcal{B}\left[k\right] \right) }
    \label{eqn:r8}
\end{equation}%
\noindent
where {$\widehat{p} \left( \mathcal{A}_{i}^{(t)}|\mathcal{B}\left[k\right] \right)$} is the normalised likelihood and 
$\mathcal{B}\left[k\right]$ indicates the $k$-th affine subspace in bag {$\mathcal{B}$}.
In order to generate the new set of particles for a new frame,
the overall likelihood for each particle is considered as the particle's weight.
The likelihoods are normalised to sum to 1 using:
\begin{equation}
  \widehat{p} \left( \mathcal{A}_{i}^{(t)}|\mathcal{B}{\left[ k \right]} \right)
  =
  \frac
  {p \left( \mathcal{A}_{i}^{(t)} | \mathcal{B}{\left[ k \right]} \right)}
  {\sum_{j=1}^{N} p \left( \mathcal{A}_{j}^{(t)} | \mathcal{B}{\left[ k \right]} \right)}
  \label{eqn:r10}
\end{equation} 
\noindent
where $N$ is the number of particles. 
The individual likelihoods are obtained using:
\begin{equation}
    p \left(  \mathcal{A}_{i}^{(t)}|\mathcal{B}{\left[ k \right]} \right)
    =
    \exp \left( \frac{-\operatorname{dist}(\mathcal{A}_{i}^{(t)},\mathcal{B}{\left[ k \right]})}{\sigma} \right)
    \label{eqn:r6}
\end{equation}%
\noindent
where $\sigma$ is a fixed parameter used to ensure that large distances result in low likelihoods.
The most likely candidate subspace is deemed to represent the best particle,
which in turn indicates the new location of the tracked object:
\begin{equation}
    \Vec{z}^{(t)}_{*} = \Vec{z}_j^{(t)},
    \quad \mbox{where} \quad j = \underset{i}{\operatorname{argmax}}~~p \left( \mathcal{A}_{i}^{(t)}|\mathcal{B} \right)
    \label{eqn:r9}
\end{equation}%
\vspace{-2ex}

\subsection{Computational Complexity}
\label{Comp_complexity}

The computational complexity of the proposed tracking framework
is dominated by generating a new affine subspace and comparing two subspaces.
The subspace generation step requires {$O(Dn^2)$} operations by performing thin SVD~\cite{brand2006}.
Computing the geodesic distance between two points on Grassmann manifold {$\mathcal{G}_{D,n}$},
requires \mbox{$O(n^3+Dn^2)$} operations for calculating the principal angles.

\section{Experiments}
\label{sec:experiments}

We evaluated the accuracy of the proposed method on eight commonly used challenging videos that have ground truth%
\footnote{The videos and the corresponding ground truth were obtained from
{\scriptsize\tt http://vision.ucsd.edu/\~{}bbabenko/project\_miltrack.html}
}
for object locations:
{\it Girl}~\cite{birchfield1998elliptical},
\mbox{\it Occluded~Face}~\cite{adam2006robust},
\mbox{\it Occluded~Face~2},
\mbox{\it Tiger~1},
\mbox{\it Tiger~2},
\mbox{\it Coke~Can},
{\it Surfer},
and
\mbox{\it Coupon~Book}~\cite{babenko2011}.
The videos contain various challenges such as object occlusions, impostor objects, pose variations, long-term appearance changes, illumination variations and non-stationary cameras.
Example frames are shown in Fig.~\ref{fig:scr_sh3}.

{\it Occluded Face} contains a face to be tracked with an occlusion challenge due to a book covering various parts of the face.
{\it Occluded Face~2} also contains a face tracking task with occlusions, but includes long-term appearance changes due to the addition of a hat.
The {\it Girl} sequence involves tracking a face with challenges such as severe pose variations and occlusion caused by another face, acting as a distractor.
{\it Tiger~1} and {\it Tiger~2} contain a moving toy with many challenges such as frequent occlusions, pose variations, fast motion (which causes motion blur) and illumination changes.
{\it Coupon~Book} contains a book being moved around, with a very similar impostor book introduced to distract the tracker.
{\it Coke~Can} contains a specular object being moved around by hand, which is subject to occlusions, fast motion as well as severe illumination variations due to a lamp.
{\it Surfer} involves tracking of the face of a surfer with many challenges such as \mbox{non-stationary} camera,
pose variations and occlusion caused by waves.

Each video is composed of 8-bit grayscale images, resized to \mbox{320 $\times$ 240} pixels.
We used normalised pixel values (between 0 and 1) as image features.
For the sake of computational efficiency in the affine subspace representation,
we resized each candidate image region to \mbox{32 $\times$ 32},
with the number of eigenvectors ($n$) and number of previous frames ($P$) set to~3 and 5, respectively.
The number of particles ($N$) is set to 100.
Furthermore, we only consider 2D translation and scaling in the motion modelling component.

Based on preliminary experiments, a bag of size \mbox{\small $K=10$} with the update rate {\small $W=5$} is used. 
For the Brownian motion covariance matrix (Section~\ref{sec:particle_filtering_framework}),
the diagonal variances corresponding to the $x$ location, $y$ location and scale
are set to $5^2$, $5^2$ and $0.01^2$, respectively.
The parameter $\sigma$ in Eqn.~(\ref{eqn:r6}) is set to $0.01$.
We have kept the parameters fixed for all videos, to deliberately avoid optimising for any specific video.
This is reflective of real-life conditions, where a tracker must work in various environments.

The source code for the proposed tracking algorithm is available at
\href{http://arma.sourceforge.net/subspacetracker/}{\footnotesize\tt \textbf{http://arma.sourceforge.net/subspacetracker/}}

\subsection{Quantitative Comparison}

Following~\cite{babenko2011}, we evaluated tracking error using the distance (in pixels) between the center of the bounding box around the tracked object and the ground truth.
The mean of the distances over the eight videos is taken as the overall tracking error.

\begin{figure}[!h]

\begin{minipage}{\columnwidth}

  \centering
  
  \includegraphics[width=0.825\textwidth]{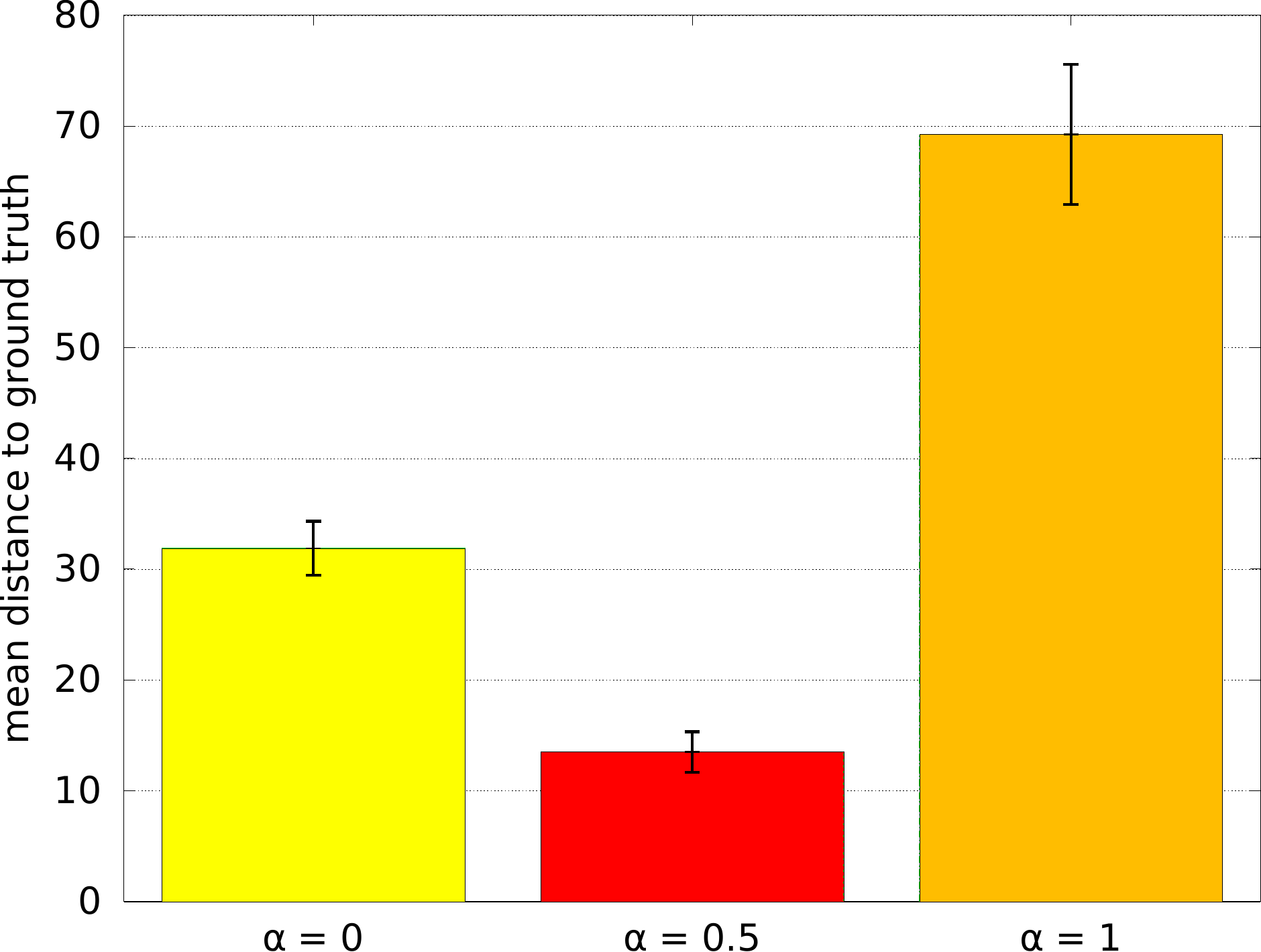}
  \vspace{-1ex}
  \caption
    {
    Tracking error for various settings of $\alpha$ in Eqn.~(\ref{eqn:r2}).
    Tracking error is measured as the distance (in pixels) between the center of the bounding box around the tracked object and the ground truth.
    For each setting of $\alpha$, the mean of the distances over the eight videos is reported.
    The bars indicate the standard error of the mean~\cite{Miller_2010}.
    \mbox{$\alpha=0$}:~only the eigenbasis is used (ie.~linear subspace),
    \mbox{$\alpha=0.5$}:~eigenbasis and mean (ie.~affine subspace),
    \mbox{$\alpha=1$}:~mean only (origins of subspaces).
    }
  \label{fig:linear}

  ~
  
  \includegraphics[width=0.825\textwidth]{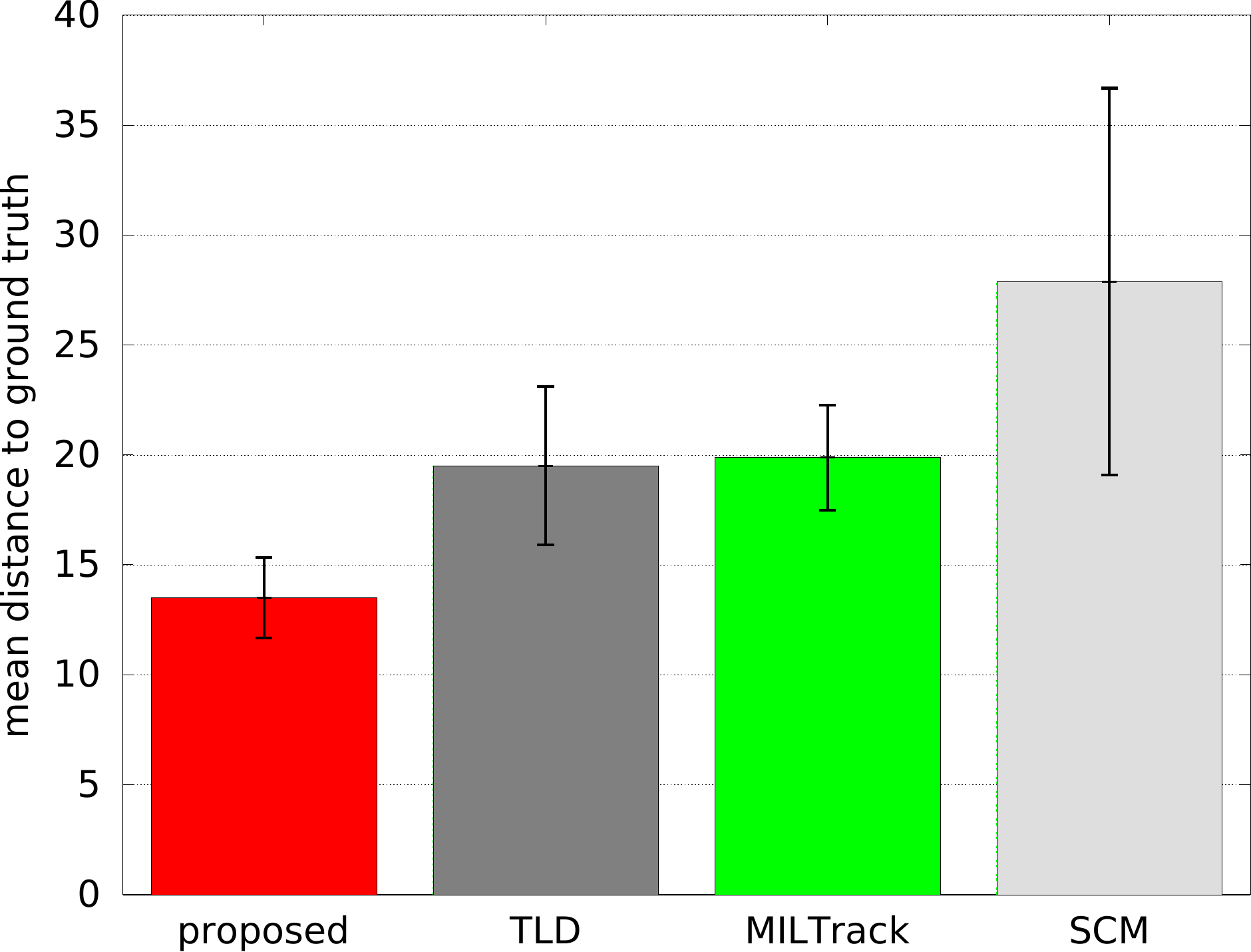}
  \vspace{-1ex}
  \caption
    {
    Comparison of the proposed method against 
    Tracking-Learning-Detection (TLD)~\cite{kalal2011tracking},
    Multiple Instance Learning Tracking~(MILTrack)~\cite{babenko2011},
    Sparsity-based Collaborative Model~(SCM)~\cite{zhong2012}.
    Tracking error is measured as per Fig.~\ref{fig:linear}.
    }
  \label{fig:errors}
  \vspace{-1ex}
\end{minipage}

\end{figure}

Fig.~\ref{fig:linear} shows the tracking error for three settings of $\alpha$ in Eqn.~(\ref{eqn:r2}).
$\alpha = 0$ ignores the origins and only uses the linear subspaces (\ie, $\Vec{\mu} = 0$ for all models);
$\alpha = 0.5$ combines the origins and subspaces;
$\alpha = 1$ uses only the origins.
Using $\alpha = 0.5$ leads to considerably lower error than the other two settings,
thereby indicating that use of the mean in conjunction with the subspace basis is effective.

\begin{figure}[!h]
  \centering
  \scalebox{1.00}{
  \begin{minipage}{1\textwidth}
    \begin{minipage}{0.025\textwidth}
      \centerline{\small (a)~}
    \end{minipage}
    \begin{minipage}{0.97\textwidth}
      \begin{minipage}{0.19\textwidth}
        \includegraphics[width=1\textwidth,keepaspectratio]{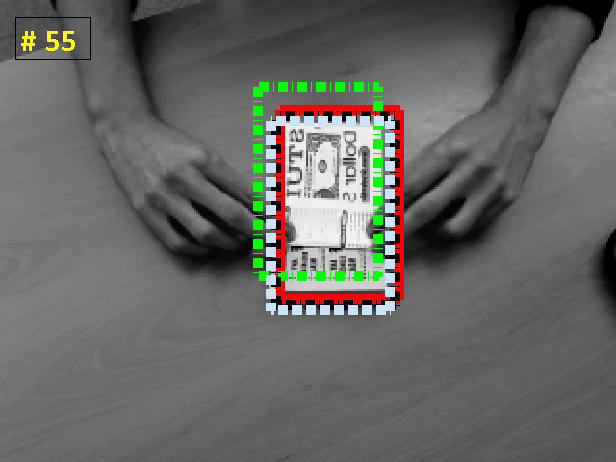}
      \end{minipage}
      \hfill
      \begin{minipage}{0.19\textwidth}
        \includegraphics[width=1\textwidth,keepaspectratio]{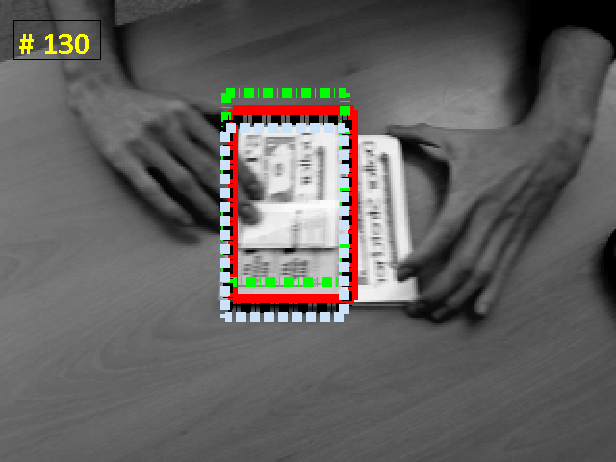}
      \end{minipage}
      \hfill
      \begin{minipage}{0.19\textwidth}
        \includegraphics[width=1\textwidth,keepaspectratio]{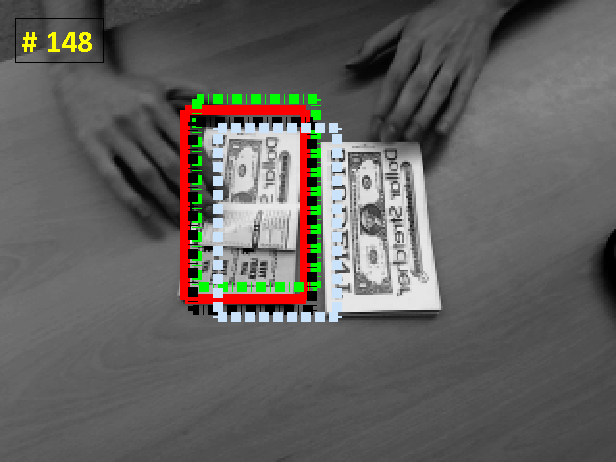}
      \end{minipage}
      \hfill
      \begin{minipage}{0.19\textwidth}
        \includegraphics[width=1\textwidth,keepaspectratio]{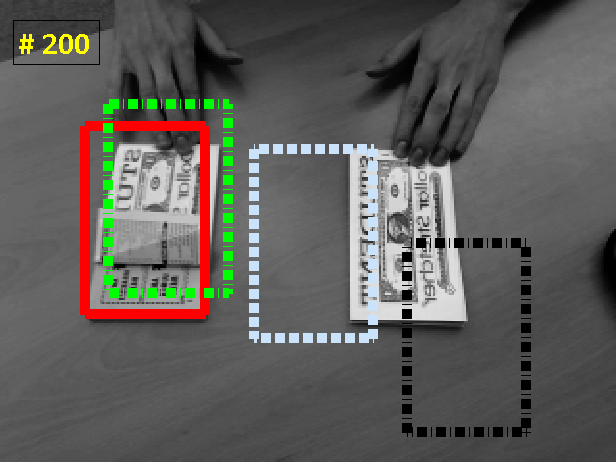}
      \end{minipage}
      \hfill
      \begin{minipage}{0.19\textwidth}
        \includegraphics[width=1\textwidth,keepaspectratio]{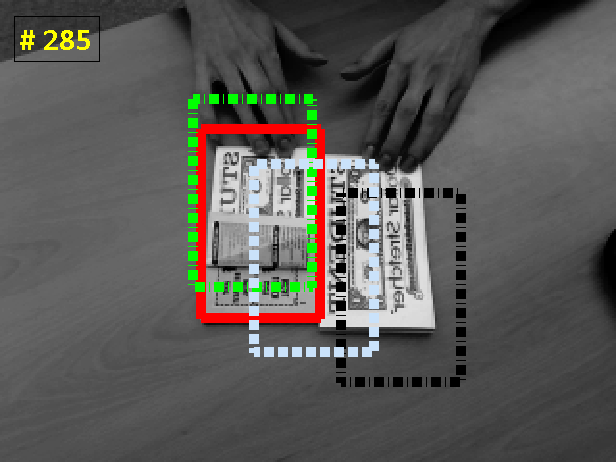}
      \end{minipage}
    \end{minipage}
  \end{minipage}
  }
  
  \vspace{1ex}
  
  \scalebox{1.00}{
  \begin{minipage}{1\textwidth}
    \begin{minipage}{0.025\textwidth}
      \centerline{\small (b)~}
    \end{minipage}
    \begin{minipage}{0.97\textwidth}
      \begin{minipage}{0.19\textwidth}
        \includegraphics[width=1\textwidth,keepaspectratio]{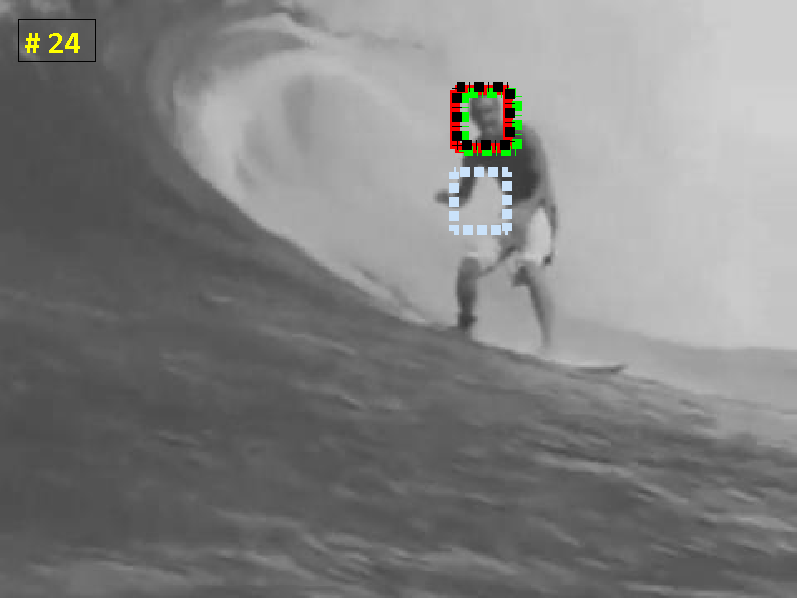}
      \end{minipage}
      \hfill
      \begin{minipage}{0.19\textwidth}
        \includegraphics[width=1\textwidth,keepaspectratio]{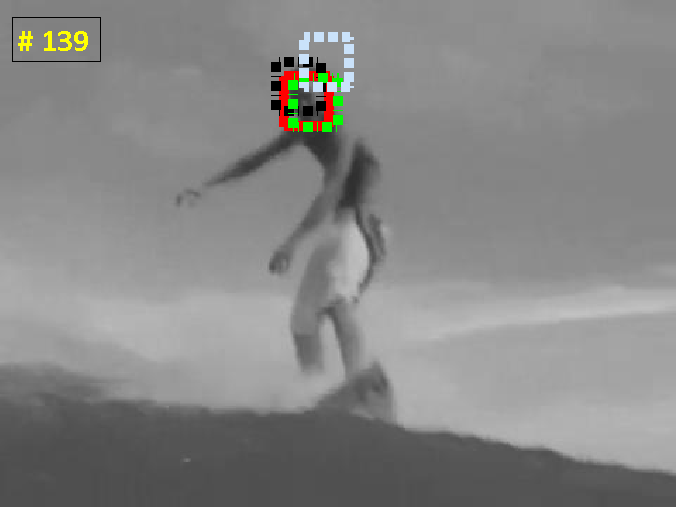}
      \end{minipage}
      \hfill
      \begin{minipage}{0.19\textwidth}
        \includegraphics[width=1\textwidth,keepaspectratio]{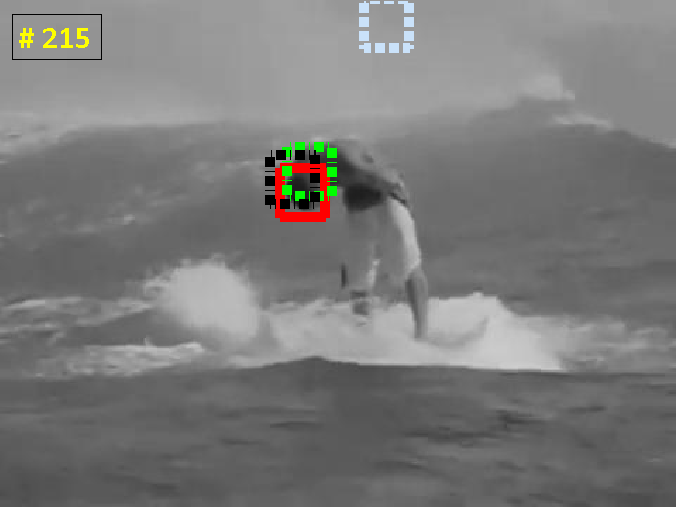}
      \end{minipage}
      \hfill
      \begin{minipage}{0.19\textwidth}
        \includegraphics[width=1\textwidth,keepaspectratio]{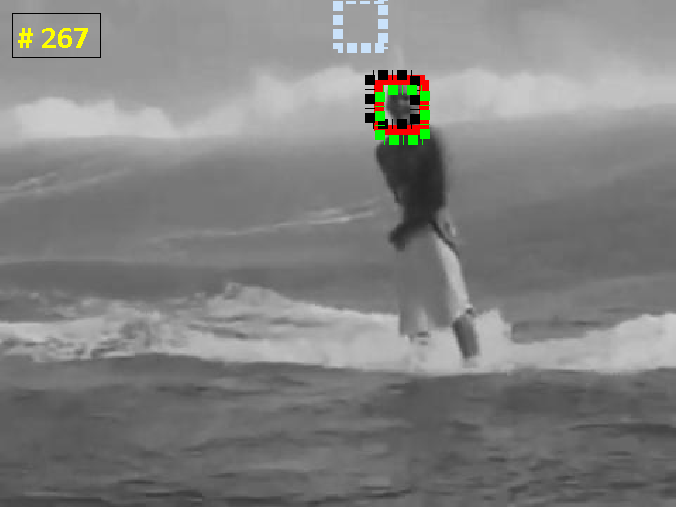}
      \end{minipage}
      \hfill
      \begin{minipage}{0.19\textwidth}
        \includegraphics[width=1\textwidth,keepaspectratio]{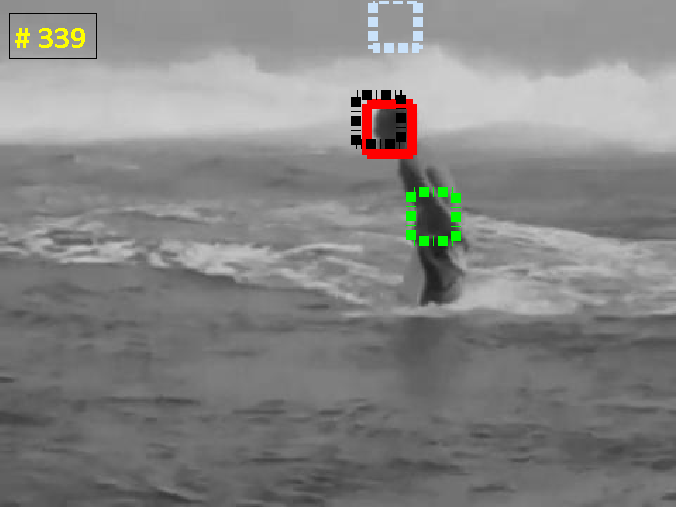}
      \end{minipage}
    \end{minipage}
  \end{minipage}
  }
  
  \vspace{1ex}
  
  \scalebox{1.00}{
  \begin{minipage}{1\textwidth}
    \begin{minipage}{0.025\textwidth}
      \centerline{\small (c)~}
    \end{minipage}
    \begin{minipage}{0.97\textwidth}
      \begin{minipage}{0.19\textwidth}
        \includegraphics[width=1\textwidth,keepaspectratio]{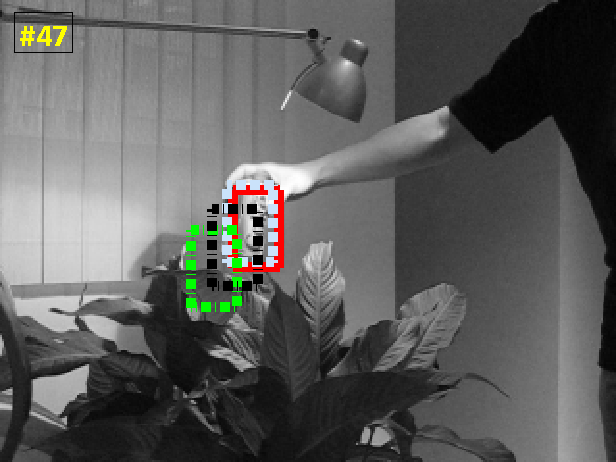}
      \end{minipage}
      \hfill
      \begin{minipage}{0.19\textwidth}
        \includegraphics[width=1\textwidth,keepaspectratio]{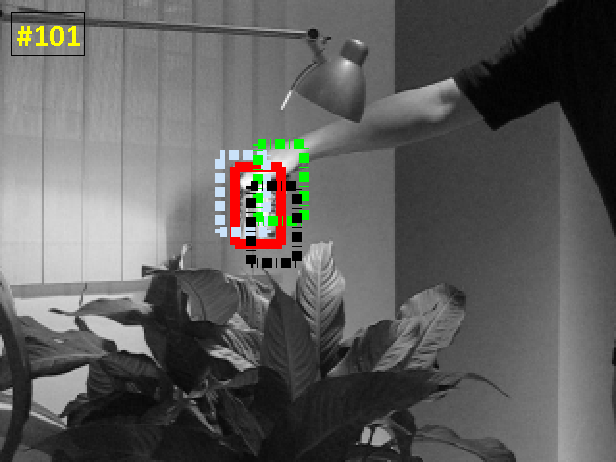}
      \end{minipage}
      \hfill
      \begin{minipage}{0.19\textwidth}
        \includegraphics[width=1\textwidth,keepaspectratio]{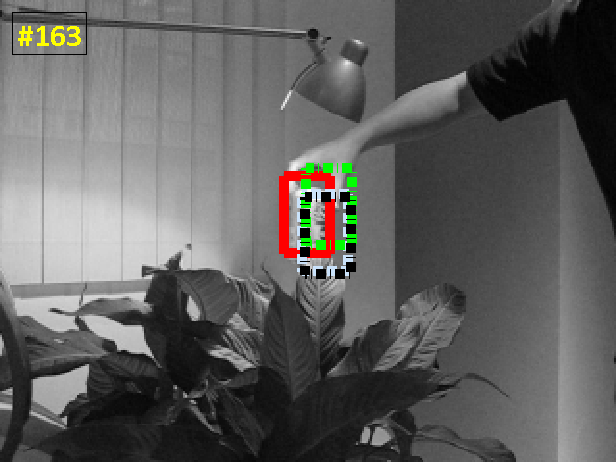}
      \end{minipage}
      \hfill
      \begin{minipage}{0.19\textwidth}
        \includegraphics[width=1\textwidth,keepaspectratio]{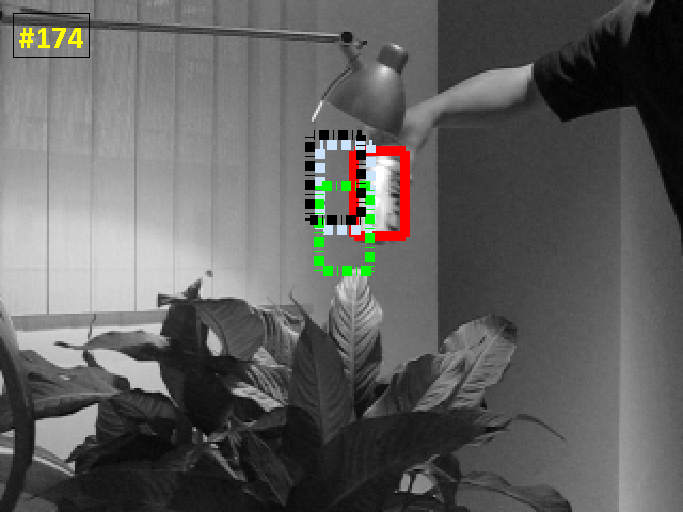}
      \end{minipage}
      \hfill
      \begin{minipage}{0.19\textwidth}
        \includegraphics[width=1\textwidth,keepaspectratio]{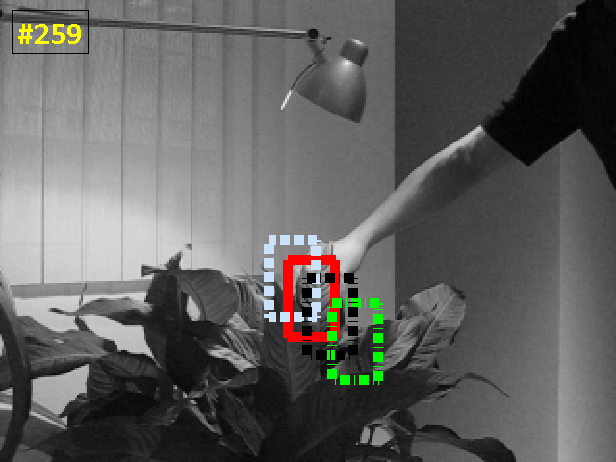}
      \end{minipage}
    \end{minipage}
  \end{minipage}
  }
  
  \vspace{1ex}
  
  \scalebox{1.00}{
  \begin{minipage}{1\textwidth}
    \begin{minipage}{0.025\textwidth}
      \centerline{\small (d)~}
    \end{minipage}
    \begin{minipage}{0.97\textwidth}
      \begin{minipage}{0.19\textwidth}
        \includegraphics[width=1\textwidth,keepaspectratio]{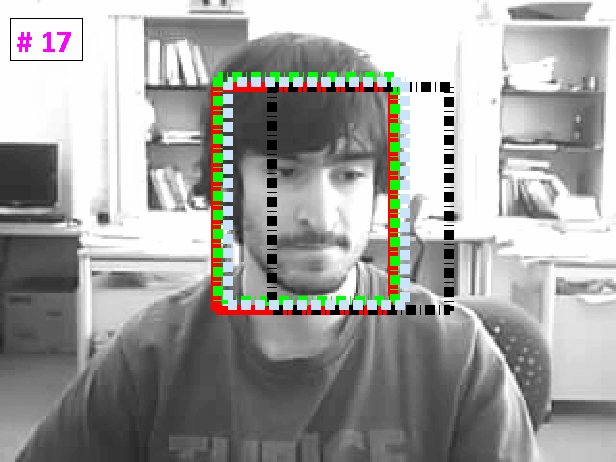}
      \end{minipage}
      \hfill
      \begin{minipage}{0.19\textwidth}
        \includegraphics[width=1\textwidth,keepaspectratio]{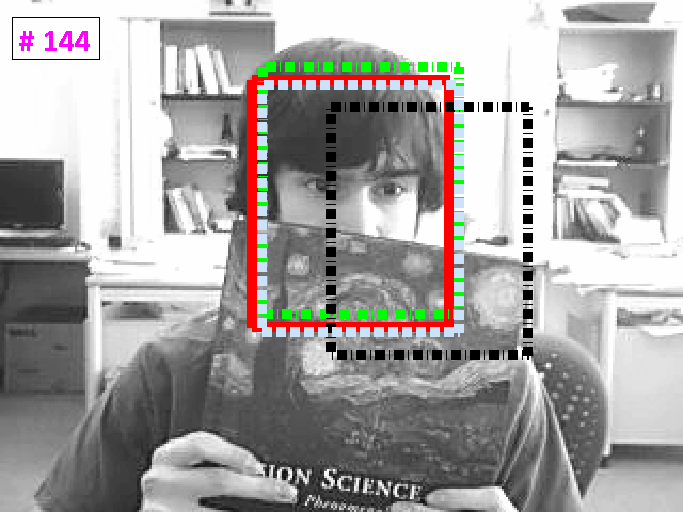}
      \end{minipage}
      \hfill
      \begin{minipage}{0.19\textwidth}
        \includegraphics[width=1\textwidth,keepaspectratio]{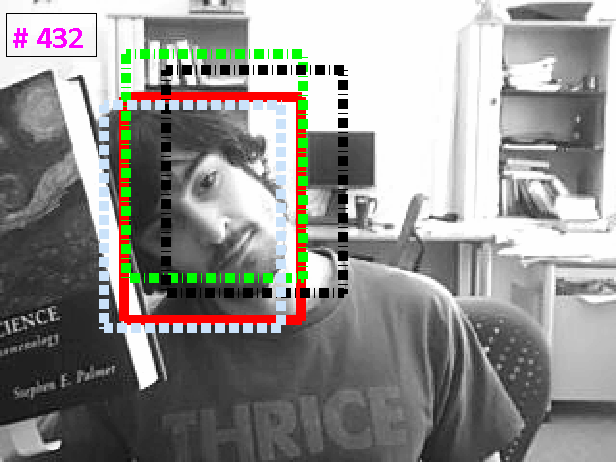}
      \end{minipage}
      \hfill
      \begin{minipage}{0.19\textwidth}
        \includegraphics[width=1\textwidth,keepaspectratio]{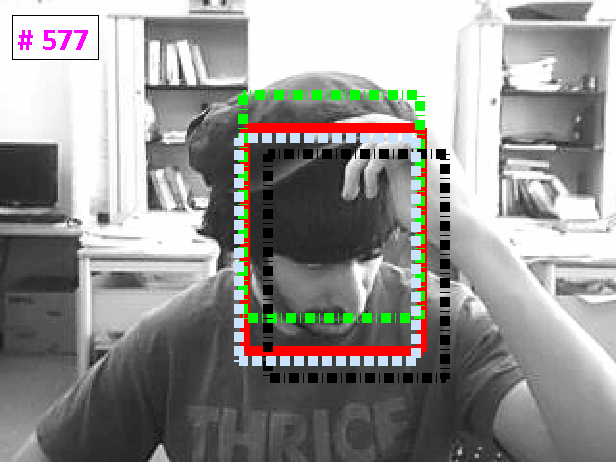}
      \end{minipage}
      \hfill
      \begin{minipage}{0.19\textwidth}
        \includegraphics[width=1\textwidth,keepaspectratio]{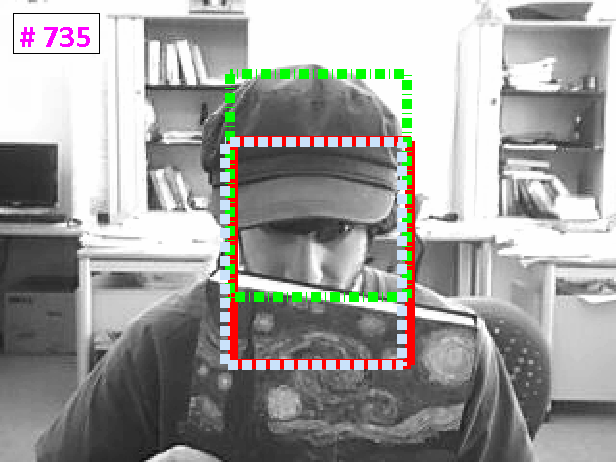}
      \end{minipage}
    \end{minipage}
  \end{minipage}
  }
  
  \vspace{1ex}
  
  \scalebox{1.00}{
  \begin{minipage}{1\textwidth}
    \begin{minipage}{0.025\textwidth}
      \centerline{\small (e)~}
    \end{minipage}
    \begin{minipage}{0.97\textwidth}
      \begin{minipage}{0.19\textwidth}
        \includegraphics[width=1\textwidth,keepaspectratio]{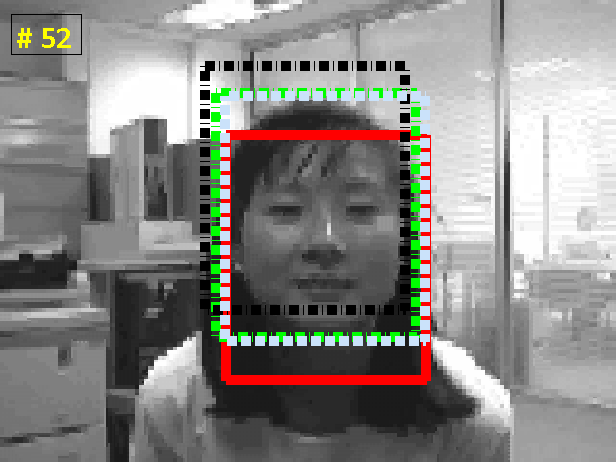}
      \end{minipage}
      \hfill
      \begin{minipage}{0.19\textwidth}
        \includegraphics[width=1\textwidth,keepaspectratio]{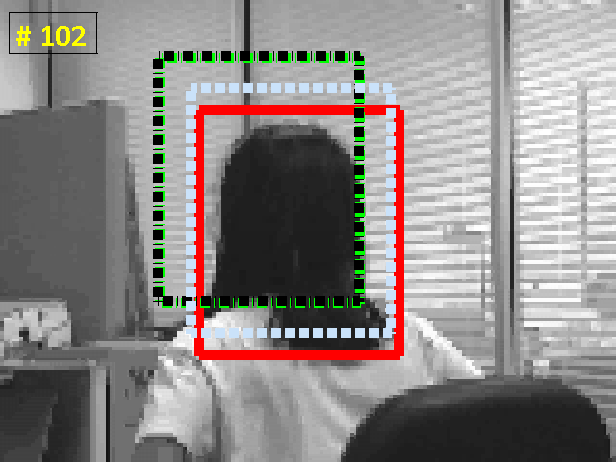}
      \end{minipage}
      \hfill
      \begin{minipage}{0.19\textwidth}
        \includegraphics[width=1\textwidth,keepaspectratio]{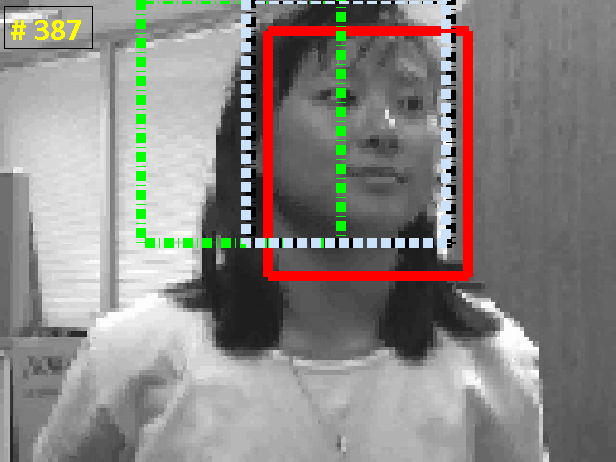}
      \end{minipage}
      \hfill
      \begin{minipage}{0.19\textwidth}
        \includegraphics[width=1\textwidth,keepaspectratio]{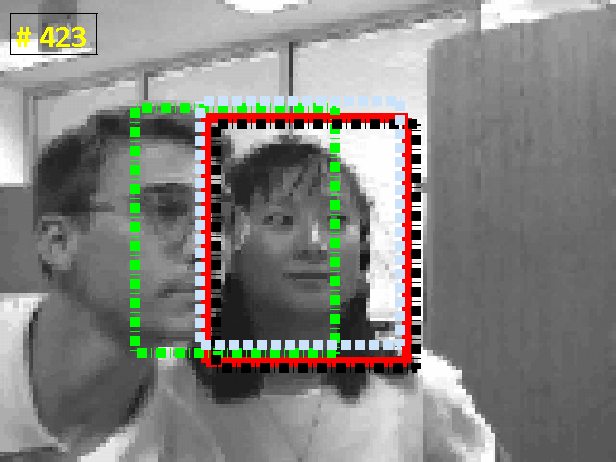}
      \end{minipage}
      \hfill
      \begin{minipage}{0.19\textwidth}
        \includegraphics[width=1\textwidth,keepaspectratio]{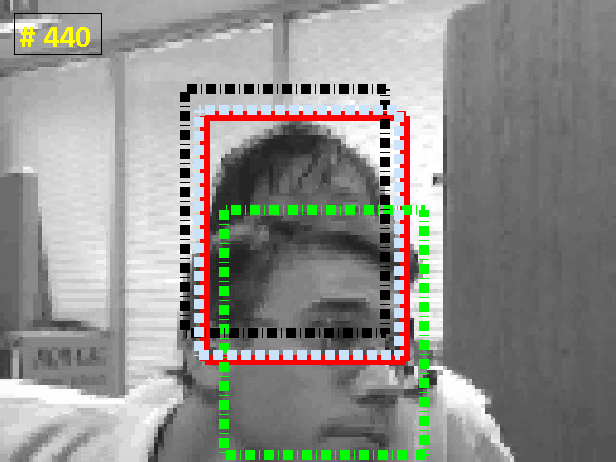}
      \end{minipage}
    \end{minipage}
  \end{minipage}
  }
  
  \vspace{2ex}
  
  \begin{minipage}{1\textwidth}
    \footnotesize
    \centering
    
    \begin{tabular}{|c|c|c|c|c|c|}
    \hline
    \multirow{2}{*}{\bf Legend:}
    & \includegraphics[width=0.06\textwidth,keepaspectratio]{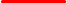}
    & \includegraphics[width=0.06\textwidth,keepaspectratio]{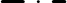}
    & \includegraphics[width=0.06\textwidth,keepaspectratio]{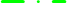}
    & \includegraphics[width=0.06\textwidth,keepaspectratio]{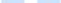}
    \\
    ~
    & proposed method
    & TLD~\cite{kalal2011tracking}
    & MILTrack~\cite{babenko2011}
    & SCM~\cite{zhong2012}\\
    \hline
    \end{tabular}
  \end{minipage}
  
  \caption
    {
    Examples of bounding boxes resulting from tracking on several videos containing occlusions, distractors/impostors, pose variations and variable object illumination.
    Best viewed in colour.
    Frames from the following videos are shown:
    {(a)}~{\it Coupon~Book},
    {(b)}~{\it Surfer},
    {(c)}~{\it Coke~Can},
    {(d)}~{\it Occluded~Face~2}~\cite{babenko2011},
    and {(e)}~{\it Girl}~\cite{birchfield1998elliptical}.
    }
  \label{fig:scr_sh3}

  \vspace{-5ex}
\end{figure}

Fig.~\ref{fig:errors} compares the tracking error of proposed tracker against three recent methods:
Tracking-Learning-Detection (TLD)~\cite{kalal2011tracking},
Multiple Instance Learning Tracker (MILTrack)~\cite{babenko2011},
and Sparsity-based Collaborative Model (SCM)~\cite{zhong2012}.
For simplicity, the proposed tracker used $\alpha = 0.5$ in Eqn.~(\ref{eqn:r2}).
Fig.~\ref{fig:scr_sh3} shows the resulting bounding boxes for several frames
from the {\it Coupon Book}, {\it Surfer}, {\it Coke Can}, {\it Occluded Face 2}, and {\it Girl} videos.
We use the publicly available source codes for
MILTrack\footnotemark[2],
TLD%
\footnote{{\scriptsize\tt http://info.ee.surrey.ac.uk/Personal/Z.Kalal/tld.html}},
and
SCM%
\footnote{{\scriptsize\tt http://ice.dlut.edu.cn/lu/Project/cvpr12\_scm/cvpr12\_scm.htm}}.

The proposed method obtains notably lower tracking error than TLD, MILTrack and SCM.
Compared to TLD (the second best tracker), the mean distance to ground truth has decreased by more than 30\%.
Furthermore, the standard error of the mean~\cite{Miller_2010} for the proposed tracker is considerably lower, indicating more consistent performance.

\onecolumn

\begin{multicols}{2}

\subsection{Qualitative Comparison}

On the {\it Coupon Book} video, TLD and SCM are confused by the distractor/impostor book.
While MILTrack mostly stays with the original book, its accuracy is lower than 
the proposed method which consistently stays centered on the original book, unaffected by the impostor book.
On the {\it Surfer} video, the proposed method and TLD consistently track the person's face.
This is in contrast to SCM which quickly loses track, and MILTrack which drifts towards the end of the video. 
On the {\it Coke Can} video, which contains dramatic illumination changes and rapid movement, MILTrack loses track after a part of the object is almost faded by the lamp light.
SCM and TLD are affected to a lesser extent.
In contrast, the proposed method consistently tracks the can, unaffected by the illumination variations.
On the {\it Occluded Face~2} video, SCM and TLD lose accuracy due to confusion by occlusions,
while SCM and the proposed method correctly track the face.
On the {\it Girl} video, the proposed method and SCM manage to track the correct person throughout the whole video.
TLD is affected by the severe pose variation (ie.~the person turning around) but recovers when the face appears frontal again.
MILTrack loses track after the pose change and then tracks the distractor/impostor face.
Overall, the qualitative observations agree with the quantitative results, with the proposed method achieving the lowest tracking error.

\section{Main Findings and Future Directions}
\label{sec:conclusion}

In this paper we addressed the problem of object tracking subject to appearance changes due to occlusions as well as variations in illumination and pose.
We proposed an adaptive tracking approach where the object is modelled as a continuously updated bag of affine subspaces,
with each subspace constructed from the object's appearance over several consecutive frames.
The bag of affine subspaces takes into account drastic appearance changes that are not well modelled by individual subspaces, such as occlusions.
Furthermore, during the search for the object's location in a new frame,
we proposed to represent the candidate image areas also as affine subspaces,
by including the immediate tracking history over several frames.
Distances between affine subspaces from the object model and candidate areas are obtained by exploiting the non-Euclidean geometry of Grassmann manifolds.
The use of bags of affine subspaces was embedded in a particle filtering framework.

Comparative evaluations on challenging videos against several recent discriminative trackers,
such as Tracking-Learning-Detection~\cite{kalal2011tracking} and Multiple Instance Learning Tracking~\cite{babenko2011},
show that the proposed approach obtains notably better accuracy and consistency.
The proposed approach also has the benefit of not requiring a separate training phase.

Future research directions include extending the bag update process to follow a semi-supervised fashion,
where the effectiveness of a new learned affine subspace is used to determine whether the subspace should be added to the bag.
Furthermore, the bag size and update rate can be dynamic, possibly dependent on the degree of tracking difficulty in challenging scenarios.


\renewcommand{\baselinestretch}{0.9}
\small\normalsize

\section*{Acknowledgements}
\footnotesize
\noindent
The Australian Centre for Robotic Vision is supported by the Australian Research Council via the Centre of Excellence program.
NICTA is funded by the Australian Government through the Department of Communications,
and the Australian Research Council through the ICT Centre of Excellence program.

\vspace{-0.5ex}

\bibliographystyle{ieee}
\bibliography{references}

\end{multicols}

\end{document}